\documentclass[preprint,12pt]{elsarticle}

\usepackage{amssymb}
\usepackage{array}
\usepackage{stfloats}
\usepackage{url}
\usepackage{amsmath}
\usepackage{caption}
\usepackage{subcaption}
\usepackage{graphicx}
\usepackage{rotating}
\usepackage{array}
\usepackage[dvipsnames]{xcolor}
\definecolor{trav}{RGB}{127, 179, 213}
\definecolor{no_trav}{RGB}{241, 148, 138}
\definecolor{fp}{RGB}{106, 90, 205}
\definecolor{fn}{RGB}{255,195,0}
\journal{Engineering Applications of Artificial Intelligence}

\begin{document}

\begin{frontmatter}



\title{Three-Dimensional Sparse Convolutional Neural Network for Terrain Traversability Analysis and Autonomous Motion Planning}


\author{Antonio Santo}
\ead{a.santo@umh.es}
\author{Juan J. Cabrera}
\ead{juan.cabreram@umh.es}
\author{David Valiente}
\ead{dvaliente@umh.es}
\author{Carlos Viegas}
\ead{carlos.viegas@uc.pt}
\author{Arturo Gil}
\ead{arturo.gil@umh.es}

\affiliation{organization={University Institute for Engineering Research, Miguel Hernández University.},
            addressline={Avda. de la Universidad s/n}, 
            city={Elche (Alicante)},
            postcode={03202}, 
            country={Spain}}

\affiliation{organization={Valencian Graduate School and Research Network of Artificial Intelligence (valgrAI)},
            addressline={Camí de Vera s/n}, 
            city={Valencia},
            postcode={46022}, 
            country={Spain}}

\affiliation{organization={Univ. of Coimbra, ADAI, Department of Mechanical Engineering},
            addressline={Rua Luís Reis Santos, Pólo II}, 
            city={Coimbra},
            postcode={3030-289}, 
            country={Portugal}}

\begin{abstract}
Among the multitude of scenarios in which autonomous robots are intended to operate, natural environments present the most significant challenges in the context of traversability estimation. Conversely, although comparatively less complex, structured environments are the most prevalent in the broader sense of autonomous navigation due to the abundance of human-designed spaces, which offer more predictable features for navigation systems. Therefore, both of these considerations lead to the conclusion that there is a need to develop methods that are specifically tailored to unstructured environments, while ensuring that the performance in structured environments is not significantly compromised. In this context, this paper makes a contribution to the state of the art with TE-NeXt (Traversability Estimation Convolutional Network). This network is a customised and efficient architecture for traversability estimation from sparse LiDAR (Light Detection and Ranging) point clouds based on a encoder-decoder topology that includes several modifications regarding: (i) the input features; (ii) the structure of encoder-decoder resolution levels; and (iii) the constitution of the 3D (three-dimensional) convolutional block. The modifications provide the robot with a deeper understanding of traversability in natural and unstructured environments. Thus, the designed architecture presents a trade-off solution, exhibiting superior performance in challenging unstructured environments (the Rellis-3D dataset), with an 82\% F1 score, which outperforms existing methodologies, while maintaining high reliability and robustness in urban environments (the SemanticKITTI dataset). Furthermore, a fully autonomous navigation framework is presented in detail. The implementation of the presented approach is available at an open repository to ensure result reproducibility.
\end{abstract}

\begin{graphicalabstract}
\centering
\includegraphics[width=1\textwidth, height=0.6\textwidth]{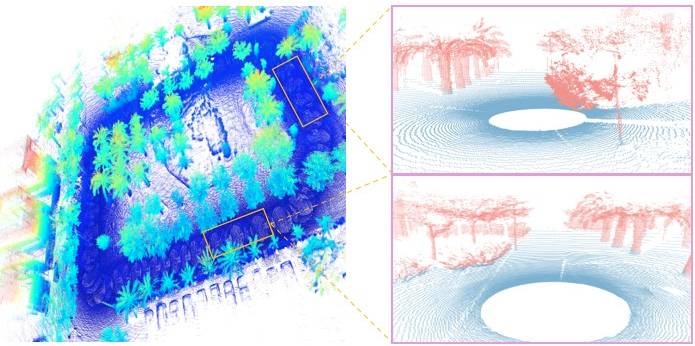}
\end{graphicalabstract}

\begin{highlights}
\item Sparse Convolutional Neural Network for traversability estimation in complex environments.
\item Novel sparse convolution block inspired by ConvNeXt, surpassing ResNet Blocks.
\item F1-5\% improvement in traversability estimation for unstructured terrain.
\item Fully autonomous navigation tool based on the proposed method on a robotic platform.
\end{highlights}


\end{frontmatter}


\section{Introduction}

In a world where autonomous navigation is a reality, the estimation of traversability has become a significant area of study in its own right. This is because it plays a crucial role in the control of autonomous vehicles in a wide range of applications.

While the estimation of traversability has now been widely acknowledged as a critical capability for mobile robots, it has historically been oriented towards the avoidance of obstacles. In this traditional context, the primary objective of the robot is to avoid any physical contact with its surroundings and to navigate exclusively through open spaces, relying on measurements from proximity sensors to detect nearby objects \cite{khatib1986real}. Subsequently, a multitude of strategies were devised to guarantee secure navigation through exteroceptive data. These included the development of alternative spatial representations, such as two-dimensional \cite{moravec1985high, hess2016real} and 2.5D maps \cite{langer1994behavior,ye2004method}.

Although this approach has demonstrated efficacy in structured contexts, the inherent limitations of these methodologies when deployed in diverse natural outdoor environments, \cite{xiao2022motion, frey2023fast}, in conjunction with the advancement of sensor technologies in terms of cost-effectiveness, precision, and portability are prompting a reassessment of the methodologies utilized to evaluate traversability in unstructured environments. The natural environments present a series of distinctive challenges, largely due to the presence of uneven terrain, pronounced textures, and irregular features. In light of these unique hazards, there is a growing necessity for autonomous vehicles to achieve a more comprehensive understanding of their surrounding environment.

The foundation for traversability estimation in unstructured environments was laid by pioneering programs such as DEMO III \cite{shoemaker1998demo}, PerceptOR \cite{kelly2006toward} and DARPA grand challenge, which encouraged the development of innovative approaches to perception, learning, and control systems for autonomous ground vehicles navigating challenging off-road terrains. The resulting methodologies established three primary approaches: pure geometric perspective, geometric and vision-based methods, and vehicle-terrain interaction-based techniques. However, the landscape of traversability estimation has evolved significantly with the emergence of deep vision technologies. This advancement has led to a decline in purely geometric approaches and highlighted limitations in methods based on vehicle-terrain interactions, particularly due to the difficulty in obtaining consistent negative examples. As a result, the field has shifted towards more sophisticated, geometric and 3D vision-based techniques that can better address the complexities of traversability estimation in natural environments.

To enhance the efficacy of specialised techniques for off-road traversability estimation, this study proposes TE-NeXt, a LiDAR-based 3D sparse convolutional architecture capable of discerning traversable and non-traversable areas captured by a LiDAR sensor, enriching the perception of the environment with semantic information in addition to geometric information.  In summary, the main contributions of this paper are:

\begin{itemize}
\item TE-NeXt: A customised 3D sparse convolutional neural network architecture for Traversability Estimation. The architecture has been specifically designed and optimised for this problem.

%

\item The design of a new convolutional block: A residual block entirely composed of 3D sparse convolutions that follows the philosophy proposed by ConvNeXt \cite{liu2022convnet} and surpasses the performance of ResNet Blocks \cite{szegedy2017inception}.

\item A real-world application of the proposed method in an autonomous navigation system in a semi-structured environment.
\end{itemize}

Consequently, the proposed network is capable of outperforming existing methods in unstructured environments while maintaining robustness and high precision ratios in urban environments.

The remaining sections of the paper are structured as follows: Section \ref{section2} outlines the prevailing methodologies employed in addressing the issue of traversability estimation. Section \ref{section3} presents the proposed architecture in detail. Following, Section \ref{section4} outlines the methodology employed to obtain the final architecture presented. Section \ref{section5} compares the proposed method with other state-of-the-art methods, demonstrating the advantages of the former. Finally, Section \ref{section6} presents the conclusions drawn from the study. 

\section{Related Work}
\label{section2}
This section provides an overview of the methods that constitute the literature on traversability estimation. It will mention the most relevant deep learning works under which the problem was originally addressed, {as well as the current methodologies employed.

\subsection{Vision-based traversability estimation} \label{subsec:vision}
The advent of deep learning methods, particularly in the domain of computer vision, has precipitated a paradigm shift. In contrast to traditional traversability estimation methods, which are based on hand-crafted features and heuristics, these new methods rely on raw sensor data, such as images or point clouds, to learn feature representations. Furthermore, they attracted considerable interest from the research community due to the more visual nature of the information obtained from the environment \cite{beycimen2023comprehensive}. Consequently, appearance-based methods became one of the most widely used approaches to this task.

One of the earliest supervised methods was proposed in \cite{selvathai2017road}, where a Multi-Layer Perceptron (MLP) classifier combined with colour and texture features (extracted from RGB images) was used as a method for automatic classification of road and off-road terrain. However, the methods rapidly evolved towards the utilisation of visual information as an input, largely due to the impressive results demonstrated by convolutional networks in image classification tasks. In this context, network architectures such as GA-Nav \cite{guan2022ga} and DeconvNet \cite{sharma2019semantic} emerged. Both solely employ visual information as an input for terrain classification, with the objective of identifying safe traversal routes. Additionally, both approaches employed semantic segmentation, grouping the results into six and five classes, respectively. The reliability of these techniques is fundamentally limited by environmental dynamics and data capture constraints during vehicle-based image acquisition \cite{sevastopoulos2022survey}.

In addition to methods that use visible spectrum images, methods have also been developed that rely on the infrared spectrum to estimate the traversability of the environment in order to avoid low-light situations such as night, backlight, haze, etc...  Pioneering works such as \cite{valada2017deep} present a convolutional neural network (CNN) for traversability estimation, and compare the optimal way to implement multi-model data fusion, now known as early or late fusion. URTSegNet \cite{liu2023urtsegnet}, inspired by methods that perform segmentation in infrared images \cite{xiong2021mcnet}, proposes a real-time neural network including two diferent branches to extract spatial detail and contextual features. Infrared imaging is significantly less effective in scenarios characterized by low thermal contrast between environmental elements. As a result, sophisticated and highly specialised calibration methods are required to derive precise measurements \cite{luo2023infrared}.

\subsection{Geometric and hybrid methods for traversability estimation} \label{3d}

The advancement of high-resolution data processing techniques \cite{graham20183d} has enabled the extrapolation of the concept of traversability, which was previously approached from a two-dimensional perspective, to a three-dimensional domain. Point clouds produced by LiDAR sensors have served as the basis for traversability estimation methods in a three-dimensional environment \cite{bae2023self,lee2021self,seo2023scate}, although pre-processing is required.

A common practice in a three dimensional context is 3D voxel partition, in approaches such as \cite{frey2022locomotion,ruetz2024foresttrav} the authors represent the point cloud as a 3D voxel occupancy map. This voxel representation avoids the need for commonly used elevation maps, which are susceptible to inaccuracies in the presence of overhanging obstacles and in multi-storey or low ceiling scenarios. In addition to the internal representation of the environment, they proposed a 3D sparse encoder-decoder that exploits the sparsity of the voxel representation for efficient computation. However, there are various methods of representing point clouds. An approach that is outlined in \cite{fusaro2023pyramidal}, involves the arrangement of point clouds within a multilevel polar grid, which naturally takes into account the non-uniform density of point clouds. An alternative approach is presented in \cite{zhou2008cylinder3d}, where point clouds are represented using cylindrical voxels based on the cylindrical coordinate system. These approaches mitigate the natural dispersion of data by creating three-dimensional maps in advance, which results in higher computational costs.

As with 2D vision, methods that fused input information of different nature were developed in the 3D context. One of the most common proposals is the merging of surface normals extracted from LiDAR data with RGB images \cite{min2022orfd,ye2023m2f2,li2024roadformer}. All these studies utilise multi-modal fusion of RGB images and geometric information derived from LiDAR point clouds. The main issue associated with the integration of different data types may be attributed to the misalignment of the timelines of the two sensors, which can lead to projection errors and generate inaccurate information.


\begin{sidewaystable}
\scriptsize
\caption{Overview of related works.}
\centering
\begin{tabular}{|>{\centering\arraybackslash}p{3.5cm}|>{\centering\arraybackslash}p{2cm}|>{\centering\arraybackslash}p{2.5cm}|>{\centering\arraybackslash}p{5.0cm}|>{\centering\arraybackslash}p{1.5cm}|}
\hline
Reference & Input data & Method & Description & Platform-specific \\
\hline
Selvathai et al., 2017 \cite{selvathai2017road} & \multicolumn{1}{c|}{RGB images}  & \multicolumn{1}{c|}{MLP} & \multicolumn{1}{l|}{\begin{tabular}[c]{@{}l@{}}Classification into road and off-road categories.\end{tabular}} & No  \\
\hline
Guan et al., 2022 \cite{guan2022ga} & \multicolumn{1}{c|}{RGB images}  & \multicolumn{1}{c|}{\begin{tabular}[c]{@{}c@{}}CNN-Attention Module \\ Deep Reinforcement Learning\end{tabular}}  & \multicolumn{1}{l|}{\begin{tabular}[c]{@{}l@{}}Segmentation of different types of terrains \\based on their navigability.\end{tabular}}  & Yes \\
\hline
Sharma et al., 2017 \cite{sharma2019semantic} & \multicolumn{1}{c|}{RGB images}  & \multicolumn{1}{c|}{CNN} & \multicolumn{1}{l|}{\begin{tabular}[c]{@{}l@{}} A semantic segmentation method for off-road \\autonomous driving using transfer learning.\end{tabular}} &  No\\
\hline
Valada et al., 2016 \cite{valada2017deep} & \multicolumn{1}{c|}{\begin{tabular}[c]{@{}c@{}}RGB, \\Infrared images \end{tabular}}  & \multicolumn{1}{c|}{CNN}  &\multicolumn{1}{l|}{\begin{tabular}[c]{@{}l@{}}Early and late fusion comparison with \\images of differing natures.\end{tabular}}  & No \\
\hline
Liu et al., 2023 \cite{liu2023urtsegnet} & \multicolumn{1}{c|}{Infrared images} & \multicolumn{1}{c|}{CNN}  & \multicolumn{1}{l|}{\begin{tabular}[c]{@{}l@{}}A semantic segmentation method for \\infrared road detection.\end{tabular}} & No  \\
\hline
Xiong et al., 2021 \cite{xiong2021mcnet}  & \multicolumn{1}{c|}{Thermal images}  & \multicolumn{1}{c|}{\begin{tabular}[c]{@{}c@{}}CNN-Attention Module\end{tabular}}  &{\begin{tabular}[c]{@{}l@{}}Thermal image semantic segmentation of \\nighttime driving scenes.\end{tabular}}  &  No\\
\hline
Bae et al., 2023 \cite{bae2023self}  & \multicolumn{1}{c|}{Point clouds}  & \multicolumn{1}{c|}{\begin{tabular}[c]{@{}c@{}}CNN-Clustering\end{tabular}}  & {\begin{tabular}[c]{@{}l@{}}Binary segmentation method for traversable \\and non-traversable regions.\end{tabular}} & No \\
\hline
Lee et al., 2021 \cite{lee2021self} & \multicolumn{1}{c|}{Elevation maps} & \multicolumn{1}{c|}{MLP} &{\begin{tabular}[c]{@{}l@{}} Method employs the extraction of terrain \\features to generate a traversability map. \end{tabular}}  & Yes  \\
\hline
Seo et al., 2023 \cite{seo2023scate} & \multicolumn{1}{c|}{Point clouds}   &\multicolumn{1}{c|}{\begin{tabular}[c]{@{}c@{}}CNN-MLP\end{tabular}} &{\begin{tabular}[c]{@{}l@{}} The method employs proprioceptive sensor\\ to infer binary and continuous traversability.\end{tabular}} & No \\
\hline
Frey et al., 2022 \cite{frey2022locomotion}  & \multicolumn{1}{c|}{3D maps} & \multicolumn{1}{c|}{Sparse CNN} & {\begin{tabular}[c]{@{}l@{}} The method compute 3D occupancy traversa-\\bility map from data generated in simulation.\end{tabular}}  & No \\
\hline
Ruetz et al., 2024 \cite{ruetz2024foresttrav} & \multicolumn{1}{c|}{3D maps} & \multicolumn{1}{c|}{Sparse CNN} &{\begin{tabular}[c]{@{}l@{}} The method compute 3D occupancy traversa-\\bility map from terrain features.\end{tabular}}  & No \\
\hline
Fusaro et al., 2023 \cite{fusaro2023pyramidal} & \multicolumn{1}{c|}{Point clouds}  & \multicolumn{1}{c|}{SVM}  &{\begin{tabular}[c]{@{}l@{}} The method operates on a pyramid-polar space \\representation to infer binary traversability.\end{tabular}}    & No  \\
\hline
Min et al., 2022 \cite{min2022orfd} & \multicolumn{1}{c|}{\begin{tabular}[c]{@{}c@{}}RGB images, \\Surface Normal \end{tabular}}  &\multicolumn{1}{c|}{\begin{tabular}[c]{@{}c@{}}CNN-Attention Module \\ MLP\end{tabular}}  &{\begin{tabular}[c]{@{}l@{}} Off-road freespace detection method on the \\RGB image plane.\end{tabular}}  & No \\
\hline
Ye et al., 2023 \cite{ye2023m2f2} &\multicolumn{1}{c|}{\begin{tabular}[c]{@{}c@{}}RGB images, \\Surface Normal \end{tabular}} &\multicolumn{1}{c|}{\begin{tabular}[c]{@{}c@{}}CNN-Attention Module \\ MLP\end{tabular}}  & {\begin{tabular}[c]{@{}l@{}} A multi-modal network for freespace detection.\end{tabular}} & No \\
\hline
Li et al., 2024 \cite{li2024roadformer} &\multicolumn{1}{c|}{\begin{tabular}[c]{@{}c@{}}RGB images, \\Surface Normal \end{tabular}} &\multicolumn{1}{c|}{\begin{tabular}[c]{@{}c@{}}CNN-Attention Module \\ MLP\end{tabular}}  &{\begin{tabular}[c]{@{}l@{}} A novel Transformer architecture for semantic
\\road scene parsing.  \end{tabular}}  &  No\\
\hline
Elnoor et al., 2024 \cite{elnoor2024pronav} &\multicolumn{1}{c|}{\begin{tabular}[c]{@{}c@{}}Proprioceptive signals\end{tabular}}  & \multicolumn{1}{c|}{PCA}  & {\begin{tabular}[c]{@{}l@{}} A method that uses proprioceptive data to \\evaluate terrain’s traversability in real time.\end{tabular}} & Yes \\
\hline
Seo et al., 2023 \cite{seo2023learning}  &\multicolumn{1}{c|}{\begin{tabular}[c]{@{}c@{}}Point clouds, \\RGB images \end{tabular}}  & \multicolumn{1}{c|}{CNN} &{\begin{tabular}[c]{@{}l@{}} Method that learns traversability only from \\self-supervisions without explicit labels.\end{tabular}}  &  No\\
\hline
Wellhausen et al., 2019 \cite{wellhausen2019should} &\multicolumn{1}{c|}{\begin{tabular}[c]{@{}c@{}}RGB images, \\ Torque Signal\end{tabular}}&\multicolumn{1}{c|}{CNN}  & {\begin{tabular}[c]{@{}l@{}} The method derives a ground reaction score, \\which serves as a measure of traversability.\end{tabular}} . & Yes\\
\hline
Seo et al., 2023 \cite{seo2023metaverse} &\multicolumn{1}{c|}{\begin{tabular}[c]{@{}c@{}}Point clouds, \\IMU \end{tabular}}& \multicolumn{1}{c|}{Sparse CNN}  &{\begin{tabular}[c]{@{}l@{}} The method predicts traversability cost map \\derived from vehicle-terrain interactions.\end{tabular}}  &Yes  \\ 
\hline
Gasparino et al., 2022 \cite{gasparino2022wayfast} &\multicolumn{1}{c|}{\begin{tabular}[c]{@{}c@{}}RGB images, \\Depth image \end{tabular}}&\multicolumn{1}{c|}{CNN}  & {\begin{tabular}[c]{@{}l@{}} Traversability
prediction in image space pro-\\vides traversability coefficient for trajectories.\end{tabular}} & Yes  \\
\hline
Gao et al., 2021 \cite{gao2021fine} &\multicolumn{1}{c|}{RGB images}  &\multicolumn{1}{c|}{\begin{tabular}[c]{@{}c@{}}CNN, \\Clustering \end{tabular}}  & {\begin{tabular}[c]{@{}l@{}}Traversability is predicted by comparing the \\features to the clustered features. \end{tabular}} & No  \\
\hline
\textbf{Ours} &\multicolumn{1}{c|}{Point clouds}  &\multicolumn{1}{c|}{\begin{tabular}[c]{@{}c@{}}Sparse CNN \end{tabular}}  & {\begin{tabular}[c]{@{}l@{}}A point-wise semantic segmentation method \\for traversability estimation.\end{tabular}} & No  \\
\hline
\end{tabular}
\label{sota}
\end{sidewaystable}

\subsection{Self-supervised traversability estimation} \label{self}

Similarly, as the primary drawback of the supervised methodology is the time-consuming annotation procedure, alternative techniques have been developed with the objective of reducing the labelling process. Consequently, methodologies such as that described in \cite{elnoor2024pronav} have been proposed. These methods utilise statistical techniques, including principal component analysis (PCA), to estimate the traversability of a given surface based on the analysis of proprioceptive sensor signals. Furthermore, methodologies such as those described in \cite{seo2023metaverse, wellhausen2019should, gasparino2022wayfast, seo2023learning} automatically generate training labels from proprioceptive data in conjunction with image or LiDAR data, where labels are generated exploiting a vehicle’s driving experience from past trajectories.

However, the inability of these methods to achieve negative examples, i.e. moving through totally untraversable areas, without compromising the structure of the vehicle or robot, has prompted the development of alternative annotation processes. In solutions such as \cite{schreiber2024w}, relative annotations for pairs of points are supplied, indicating which point in the pair is more traversable, or if both points possess equal traversability. Alternative approaches, such as referenced in \cite{gao2021fine}, employ sparse anchor patches to denote regions with distinct semantic attributes regarding their traversability and classify them into existing semantic clusters.

\subsection{Current proposal} \label{hola}

This paper presents TE-NeXt, an encoder-decoder architecture optimised for traversability estimation in unstructured environments. The approach presented here can be classified as a supervised method, in accordance with the paradigm described in Subsection \ref{3d}, as it is a supervised paradigm method. However, it employs local point clouds, a distinctive feature that sets it apart from other methods, such as those described in \cite{frey2022locomotion, ruetz2024foresttrav}, which utilise point cloud maps and preprocess the data.  Moreover, in contrast to the techniques outlined in Subsection \ref{subsec:vision}, TE-NeXt employs an optical sensor that is invariant to changes in light conditions (LiDAR sensor). Furthermore, the proposed methodology is designed to overcome the limitations associated with self-supervised learning approaches. Unlike most of the existing methods, it does not require a specific platform, making it a more general and secure solution for all types of vehicles. Table \ref{sota} outlines the distinctive features of the proposed approach in comparison to existing, state-of-the-art methodologies.

The proposed architecture requires a point cloud as input, as some of the methods mentioned in this section, and the feature extraction at different scales is computed at the residual block, which inherits design notions from two broad paradigms: vision Transformers and ConvNext \cite{liu2022convnet}. The primary objective of TE-NeXt is to demonstrate enhanced performance in natural environments in comparison to existing state-of-the-art methodologies for semantic segmentation and traversability estimation. Additionally, the system is designed to operate as a viable alternative in structured environments, although this is not the ultimate intended outcome of the study.

\begin{figure}
\centering 
\includegraphics[width=1.05\textwidth, height=0.9\textwidth] {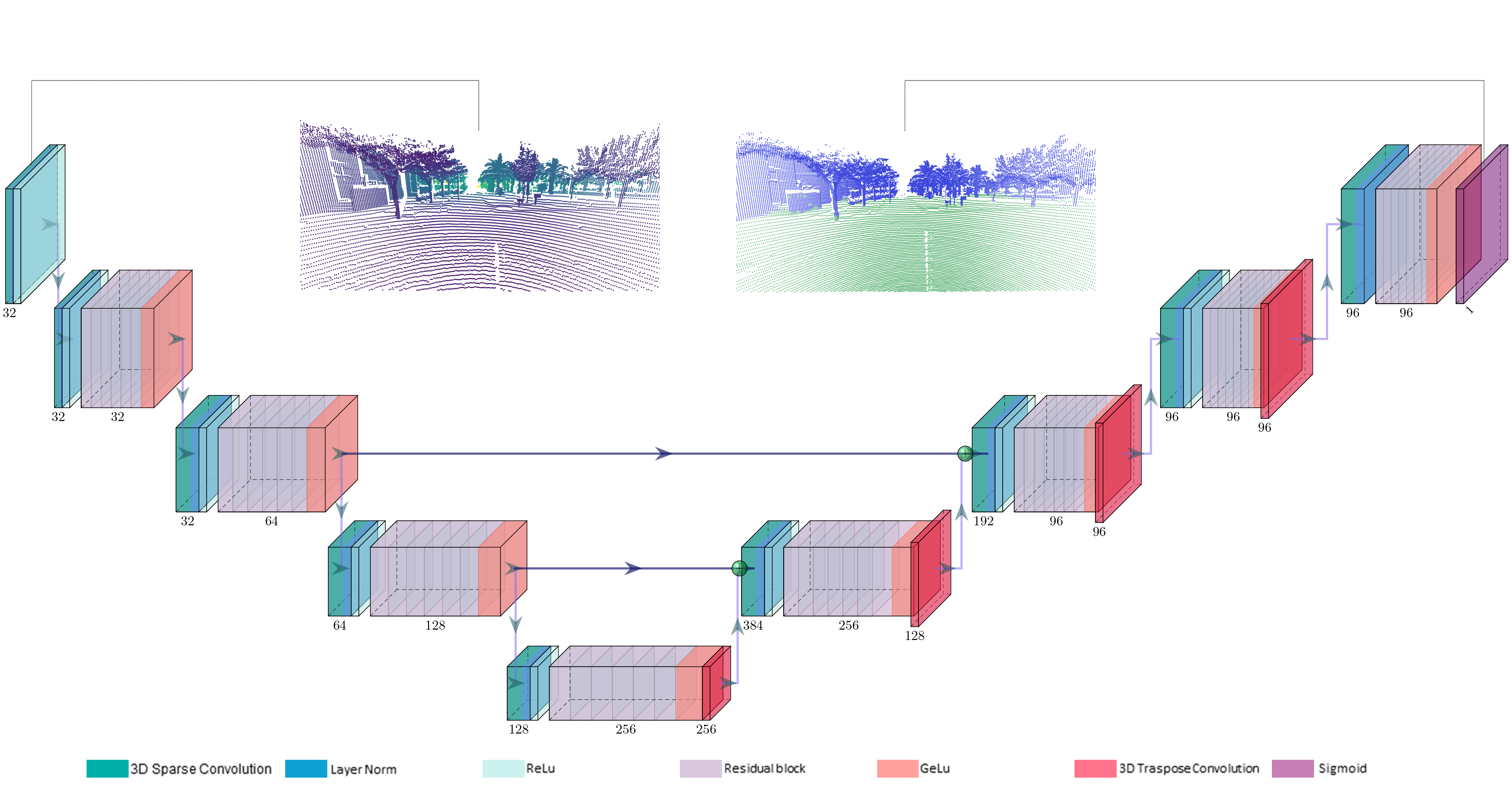}
\caption{Traversability Estimation framework overview. A detailed illustration of the proposed TE-NeXt architecture, following the original scheme of the U-Net network \cite{ronneberger2015u}. A LiDAR point cloud is subjected to the extraction of descriptors at varying resolutions on the left-hand side of the image. Once the information has been encoded, the initial dimensions are recovered in order to obtain a label for each point.}
\label{fig:TE-NeXt} 
\end{figure}

\section{TE-NeXt: AN ARCHITECTURE FOR TRAVERSABILITY ESTIMATION}
\label{section3}
TE-NeXt is a dedicated convolutional neural network, specifically designed for the task of 3D semantic segmentation of point clouds, with a particular emphasis on the estimation of traversability. TE-NeXt builds upon the U-Net style architecture \cite{ronneberger2015u}, replacing the standard residual blocks, introduced by classical ResNet networks \cite{he2016deep}, with a propietary block inspired by the ConvNeXt \cite{liu2022convnet} design principles. What sets TE-NeXt apart from recent work in this domain is its aim to learn a more general, high-level understanding of the environment that is not tied to specific settings or robotic platforms. 

\subsection{Mathematical Formulation}

The traversability estimation problem is formulated as a semantic segmentation task, where a point cloud $\mathcal{P}$ with $N$ points is characterized by Cartesian coordinates $\vec{p_i} = (x_i, y_i, z_i) \in \mathbb{R}^3$ and feature vectors $\vec{f_i} \in \mathbb{R}^{d_{\text{in}}}$. The objective is to learn a mapping function $f: \mathbb{R}^{3+d_{\text{in}}} \rightarrow [0,1]$ that assigns a traversability probability $\hat{l}_i$ to each point, where $\hat{l}_i \in [0,1]$. By training on a labeled dataset, $\{(\vec{p}_i, \vec{f}_i), \hat{l}_i\}$, the model aims to capture intricate spatial and feature-based patterns, enabling precise prediction of point-wise traversability through a deep learning approach that minimizes classification error across the entire point cloud representation.

\subsection{Network Architecture Overview}
Traversability estimation can be considered a semantic segmentation problem. Considering a simple point cloud as input, discriminant features could be extracted by preprocessing the data. The income's internal representation follows the methodology used by 3D sparse convolutional library, Minkowski Engine \cite{choy20194d}. It is defined by $I=\{(\vec{T}_c, \vec{T}_f), l_i\}$, where $\vec{T}_c=(x_i, y_i, z_i)$ comprises those coordinates provided by the collimated beams of the LiDAR sensor, ignoring the spaces where there is no information based on sparse matrix principles, whereas $\vec{T}_f$ describes point-wise features and $\l_i$ indicates the traversability condition.

As previously stated, the focus of recent works is on the identification of environmental attributes that facilitate the discernment of traversable from non-traversable areas. 
However, in other domains such as large-scale place recognition, which a priori generalisation is a fundamental requirement, the features are initiated as a vector of 1's for those non-empty voxels \cite{cabrera2024minkunext,komorowski2021minkloc3d}.  According to this view, the method presented in this paper continues such approach as it aims to optimise the abstraction process.

Next, we provide a description of the TE-NeXt architecture, depicted in Figure \ref{fig:TE-NeXt}. The proposed network adopts a hierarchical structure analogous to the U-Net \cite{ronneberger2015u}, which is an encoder-decoder configuration, with the spatial dimensions initially reduced through strided convolution operations. This enlarges the receptive field size of the neurons at deeper layers, enabling them to capture a more extensive contextual information from the input. 
Following each of these operations, a normalisation layer, LayerNorm, and an activation function, ReLu, are employed for each level of network resolution. The latter is a common technique used in convolutional networks to learn complex, non-linear mappings between inputs and outputs. Conversely, the utilisation of a normalisation technique that is not commonly employed in classical convolutional networks is justified by the promising outcomes observed in vision Transformers. The principal reason for its efficacy is its invariance with respect to the batch size, given that it is not typically feasible to process substantial quantities of data in a three-dimensional domain concurrently.

Subsequently, the spatial dimensions are incrementally augmented through upsampling operations, such as transposed convolutions. To compensate for the loss of spatial details during encoding, skip connections are employed to concatenate feature maps from the encoder path with those in the decoder path at the same resolution level. Unlike U-Net and applying the same policy of \cite{frey2022locomotion} reducing the number of skip connections to only the lower layers, it helps reduce false positive predictions, likely because the decoder does not have to predict many forwarded features from the encoder. A reduction in the number of skip connections appears to work as a form of regularisation, thus enhancing overall traversability prediction performance.

As a result, the neurons in the final layers of the decoder have a large effective receptive field, capturing both global context and local details. Finally, as a decoder, the network output holds the same dimensions as the input data, providing a probability value obtained via a Sigmoid function as the final layer. 
\subsection{TE-NeXt residual block}

\begin{figure}
\centering 
\includegraphics[width=0.65\textwidth, height=0.55\textwidth] {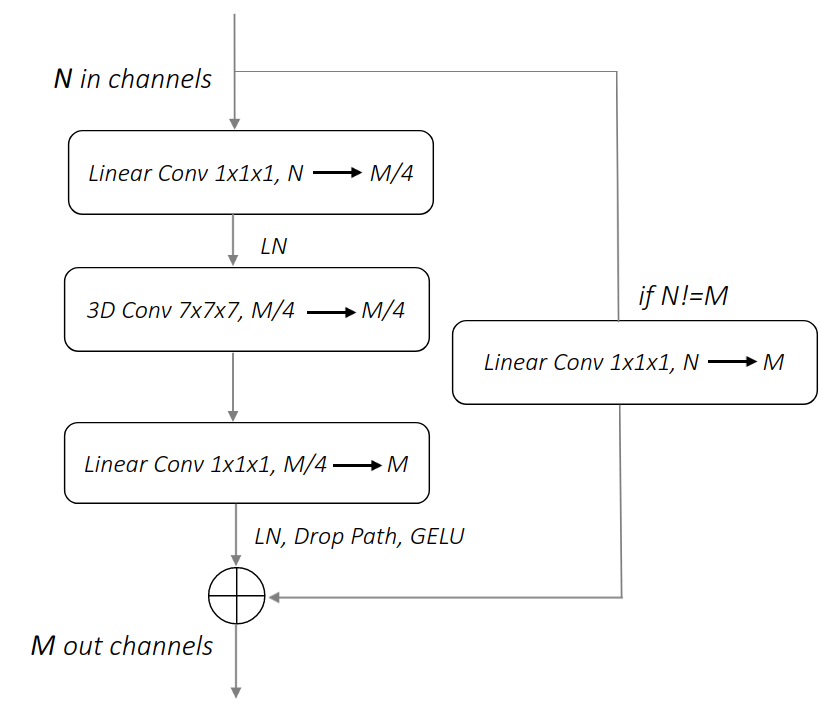}
\caption{The structure of the proposed residual block TE-NeXt.}
\label{fig2}
\end{figure}

A major innovation of TE-NeXt can be attributed to its residual block. This block is inspired by the design principles of ConvNeXt \cite{liu2022convnet}, a convolutional network architecture that has informed numerous methods which constitute the current state-of-the-art on free space detection in images \cite{li2024roadformer}. It incorporates several key ideas from Vision Transformers while maintaining the simplicity and efficiency of standard Convolutional Neural Networks (CNNs). TE-NeXt aims to learn rich, generalisable feature representations from large-scale 3D point cloud data. The residual block is illustrated in Figure \ref{fig2}.

The \textit{N}-channel input first passes through a 1x1x1 convolutional layer that reduces the channels by a factor of 4, aleviating the computational complexity and memory requirements of subsequent layers. The idea behind this operation is to facilitate the independent examination of cross-channel correlations and spatial correlations. Therefore, it operates on each channel independently, allowing channel-specific filters and weights to be learned. The 1x1x1 convolution provides a sophisticated mechanism for neural networks to simultaneously reduce computational complexity and mitigate inherent spatial and channel redundancies prevalent in standard convolutional architectures, enabling more efficient feature representation through intelligent channel-wise transformations and selective information preservation \cite{li2023scconv}. The underlying premise is that cross-channel and spatial correlations are sufficiently decoupled that mapping them together is not the optimal approach \cite{chollet2017xception}. Once the aforementioned mapping has been completed, the spatial correlations are mapped using regular 3D convolutions with a 7x7x7 kernel.

Next, another linear convolution layer increases the channels back to the desired output. In essence, the final 1x1x1 convolution allows the residual block to: i) restore the original channel dimensionality after the bottleneck layers; ii) increase representational power and iii) perform projection mapping. 
Finally, if the input and output dimensions are the equal (\textit{N}=\textit{M}), the input is added to the output via a skip connection before the normalization layer, the regularization technique applied (drop Path  \cite{DBLP:journals/corr/LarssonMS16a}) and a  Gaussian Error Linear Unit (GeLU) \cite{DBLP:journals/corr/HendrycksG16}. Regarding the activation function employed, GeLU is smoother than Rectified Linear Unit activation (ReLU) and is utilized in the most advanced Transformers. This skip connection allows gradients to flow more easily during backpropagation, mitigating the vanishing gradient problem in very deep networks. Furthermore, the residual block can learn an identity mapping when needed, making it easier to optimize.

\section{Experiments}
\label{section4}
\subsection{Datasets}

A set of publicly available databases has been used for training, validation and testing of the neural network design, in particular:
\begin{itemize}
    \item SemanticKITTI \cite{behley2021ijrr}: A large-scale dataset for semantic scene understanding of LiDAR sequences. It is based on the KITTI Vision Odometry Benchmark \cite{geiger2012we} and provides dense point-wise annotations for the complete 360° field-of-view of the Velodyne HDL-64E LiDAR sensor used in the KITTI dataset. It consists of over 43,000 LiDAR scans from 22 sequences of the KITTI Odometry Benchmark, with over 4 billion points in total. The sequences cover a variety of highly structured environments including city streets, residential areas, highways, and countryside roads.

    \item Rellis-3D \cite{jiang2021rellis}: Multimodal dataset designed for off-road robotics and autonomous navigation research, serving as an extraordinary complement to other existing autonomous driving datasets that focus on urban scenes. It consists of 13,556 point clouds divided into 4 distinct sequences captured by means of an OS1-64 LiDAR.
    
\end{itemize}

The datasets contain around 25 to different semantic labels that a point can be assigned, such as bush, mud, asphalt, sidewalk, etc. The condition labels were reclassified based on its traversability, resulting in only two classes: ``traversable'' and ``non-traversable''. The classes converted to ``traversable'' include sidewalk, asphalt, low vegetation or grass, dirt, cement, and mud. The ``non-traversable'' class comprises tree, person, car, truck, building, and others. This binary categorization was done by considering the ability of a robot to safely navigate across the different surface types.

\subsection{Training and evaluation process}

Given the considerable discrepancy in the quantity of labelled data available from urban and natural environments, being the latter the primary focus of this research, the objective was to achieve a more balanced distribution of data across these domains. In particular, 10,700 clouds of unstructured environments have been selected, belonging to four of the five sequences in Rellis-3D. Additionally, the initial four SemanticKITTI sequences have been selected, thus ensuring 11,100 clouds of structured environments. Moreover, the validation and testing process has been conducted with previously unseen sequences from both datasets to evaluate the neural network's generalisation capabilities.

\begin{table}[h]
\footnotesize
\caption{Hyperparameters selected in training process.}
\begin{center}
\begin{tabular}{lllll}
\cline{1-2}
\textbf{Config}         & \textbf{Value} &  &  &  \\ \cline{1-2}
Optimizer               &     AdamW           &  &  &  \\
Scheduler               &     Cosine           &  &  &  \\
Criteria                &     Binary Cross Entropy           &  &  &  \\
Learning Rate           &     5e-3           &  &  &  \\
Weight decay            &     0.05           &  &  &  \\
Batch size              &     5           &  &  &  \\
Datasets                &     Sem.KITTI, RELLIS-3D           &  &  &  \\
Quantization scale      &     0.2               &  &  &  \\
Warmup epochs           &     80           &  &  &  \\
Termination criteria &        F1 score       &  &  &  \\ \cline{1-2}
\end{tabular}
\end{center}
\label{Table:setup}
\end{table}

Table \ref{Table:setup} specifies the hyperparameters and settings used to train the model from randomly initialized weights. The optimizer used AdamW with an initial Learning Rate (LR) of 5e-3 along to Cosine Annealing scheduler causes the LR to follow a cosine function over the same training epoch. It performs periodic ``warm restarts'' where the learning rate is reset back to the initial value after a certain number of batches. This simulates the restarting of the learning process using the weights from the previous run as a good starting point. The periodic restarts help the model escape from local minima and saddle points. Since the model has been trained for a binary segmentation task, the F1 score is a suitable metric for imbalanced datasets and considers both: precision (the proportion of true positive instances among all positive predictions) and recall (the proportion of true positive predictions among all actual positive instances). This guarantees that the model is not optimising for one metric at the expense of the other. Moreover, should the F1 score on the validation set cease to improve or begin to decline, training can be terminated prematurely to avert overfitting.

\subsection{Ablation study}
\label{ablation_section}
This section presents the roadmap followed during the development of the architecture proposed in this paper, with the objective of adapting the initial architectural approach used for indoor semantic segmentation, such as MinkU-Net34C, to outdoor environments, particularly natural environments. For this purpose, a first distinction will be made in three blocks.

The initial study examines the impact of input features and their spatial distribution on the performance of the neural network.
The second one will present the design steps that affect the structure of the neural network. Subsequently, the methodology employed in the design of the residual block, which represents the primary innovation of this work, will be explained.


Figure \ref{ablation_steps} provides a graphical representation of the research process. During the development of the architecture, we have endeavoured to conduct experiments that are mutually related, thereby establishing a temporal and logical consistency in the nodes that comprise the graph. Consequently, the X-axis represents the design and research process, while the Y-axis represents the performance of the various models  on the Rellis-3D dataset, in terms of one metric considered relevant to the task of traversability estimation: the F1 score. This metric is regarded as a valuable measure of model performance, particularly in situations where data sets are imbalanced. It offers a well-balanced assessment of a model's precision and recall, making it an essential tool for evaluating model performance across a wide range of applications.
Precision denotes the model's capacity to avoid misclassifying non-traversable regions as traversable (i.e., preventing false positives), which is crucial for ensuring safety. Recall, on the other hand, represents the model's ability to identify all traversable regions (i.e., avoiding false negatives), which is essential for efficient navigation. The F1 metric integrates precision and recall into a unified score, enabling an assessment of the equilibrium between the two.

To facilitate the comprehension of the ablation study, Table \ref{ablation_steps_legend_2} has been introduced as a legend, providing information on the starting point of each experiment and the changes made. Additionally, the number of parameters of each model is included, revealing a notable reduction in the number of parameters compared to the starting point. A closer examination reveals that the initial block of experiments has no discernible effect on the network’s trainable parameters. However, in subsequent blocks of experiments, notable alterations in the model size emerge. These changes are due to modifications in the convolution operations, kernel sizes, and the number of parallel branches in the convolutional blocks.

\begin{figure}[t] 
\includegraphics[width=1.22\textwidth, height=0.7\textwidth]{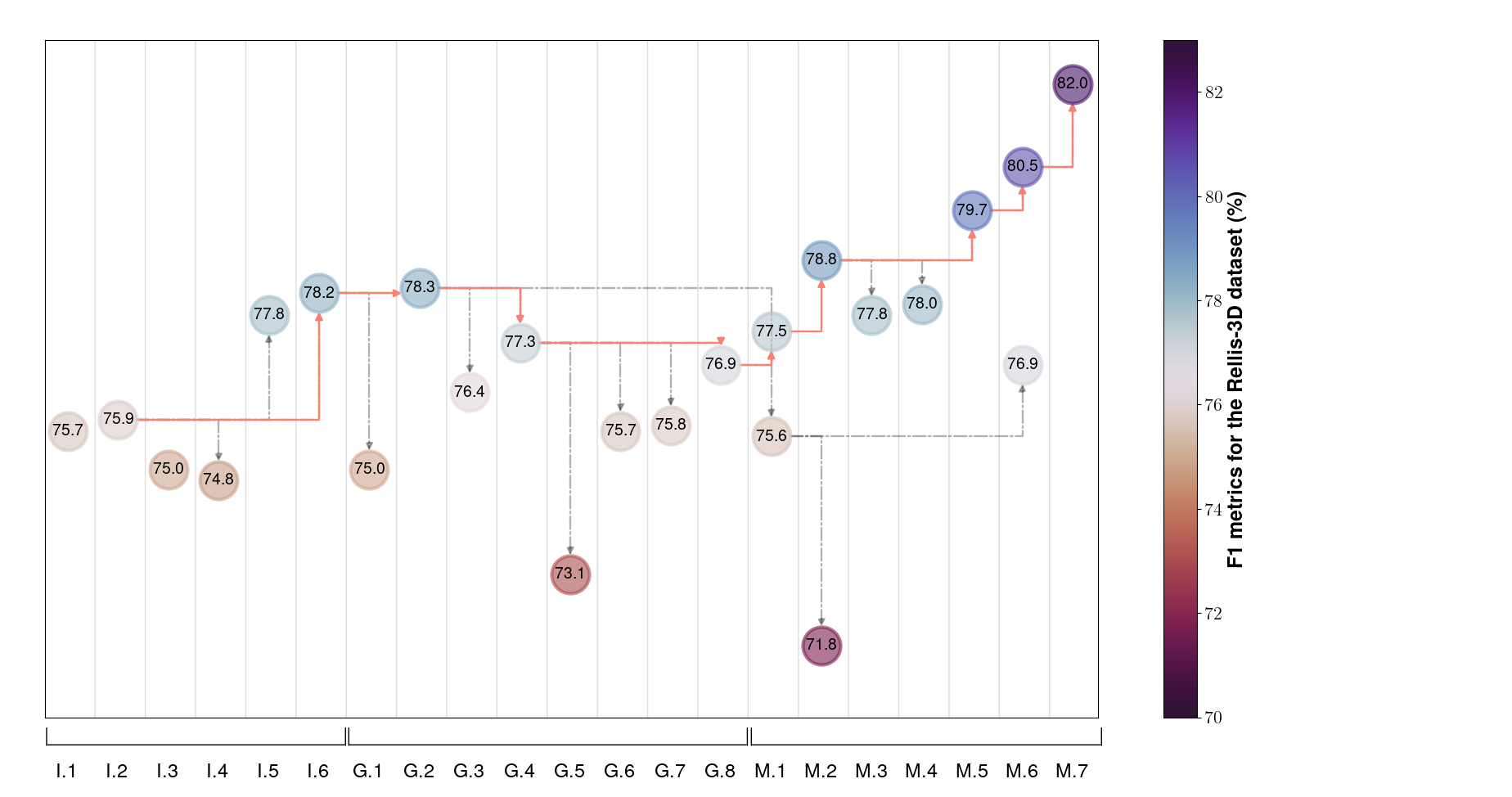}
\caption{Design process of the proposed architecture, TE-NeXt, starting from a base MinkU-Net34C. All experiments are detailed in Table \ref{ablation_steps_legend_2}.}
\label{ablation_steps}
\end{figure}

\vspace {4.1pt}
\subsubsection{Input data}
\begin{list}{}{}
    \item{\textbf{Data Discretisation Size}: Discretization involves converting continuous spatial data into discrete coordinates, storing them efficiently. This process is essential for handling high-dimensional sparse tensors in 3D data processing tasks. Nevertheless, for tasks involving three-dimensional perception, the impact of quantization on performance is relatively insignificant when the quantization size is less than 5 cm. Thus, the experiments designated as \textsl{I.1, I.2} and \textsl{I.3} have been identified as a preliminary phase in a series of subsequent experiments. 
     A finer discretization captures more information but can be computationally expensive, while coarser discretization simplifies the data at the risk of losing important features. The choice of discretization parameter should balance these factors based on the specific requirements of the task and available resources.}

    \item{\textbf{Features}}: In contrast to 2D convolutions, where the input and output tensors have a fixed number of feature dimensions, typically called channels (e.g. RGB channels for images), 3D sparse convolutions allow sparse tensors to have an arbitrary number of feature dimensions. Each non-zero element in the sparse tensor can have a feature vector of arbitrary length. Therefore, a study of the most discriminating features for estimating traversability was carried out. Among the most representative are those shown in Figure \ref{ablation_steps} (experiments \textsl{I.4, I.5}, and \textsl{I.6}, referred to in Table \ref{ablation_steps_legend_2}). Nevertheless, the vector of 1's  (initialising the feature associated with each point to a value of 1) was selected as the primary candidate for future modifications, as it is equally discriminant and, at the time of inference, represents an improvement in efficiency.
\end{list}

\subsubsection{Global Design}

\begin{list}{}{}
    \item{\textbf{Cardinality}}: This term was introduced by ResNext \cite{DBLP:journals/corr/XieGDTH16}, and refers to the number of parallel branches within each block. Related to this, a number of different configurations have been assessed. The initial configuration employed was the MinkU-Net34 network, comprising (2, 3, 4, 6, 2, 2, 2, 2, 2) branches per block. In order to reduce the number of model parameters, it was decided to test the MinkU-Net18 and MinkU-Net14, experiments \textsl{G.1} and \textsl{G.2}. The smaller versions, in particular MinkU-Net14, exhibit a notable reduction in the number of branches per convolutional block in comparison to MinkU-Net34. This leads to a decrease in the number of filters and, consequently, in the number of trainable parameters in the model.
    In conclusion, through a meticulous examination of smaller architectural configurations, highly efficient MinkU-Net configurations have been identified that strike a balance between model size and performance. This illustrates that it is frequently feasible to significantly reduce the complexity of neural networks without compromising their capacity to learn and generalise. While this modification does not lead to an significant enhancement in performance, it does contribute to greater efficiency, as will be demonstrated in Table \ref{efficiency}.

    \item{\textbf{Skip connections}}: Unlike the U-Net architecture \cite{ronneberger2015u}, Te-Next presents a variation in the connections that are made between the respective parts of the encoder and decoder. In addition to the four skip connections that were previously implemented in the experiments, two and three connections corresponding to experiments \textsl{G.3} and \textsl{G.4} have been evaluated. Notwithstanding the fact that both exhibited a reduction of 2\% in terms of the F1 metric, the more robust performance options, experiment \textsl{G.2} (later mentioned as the alternative experimental branch) and experiment \textsl{G.4} (a configuration comprising just two connections), were subjected to further investigation in subsequent experiments. This has been influenced by concepts from related research \cite{frey2022locomotion, ruetz2024foresttrav}.


    \item{\textbf{Modifications in the residual block}}: 
     The aforementioned configuration settings are based on the original convolution block of ResNetv1 \cite{DBLP:journals/corr/HeZRS15}.  Consequently, the experiments between \textsl{G.5} and \textsl{G.8} seek to examine the efficacy of alternative convolution blocks that are commonly employed in CNNs for semantic segmentation:
     \begin{itemize}
         \item Squeeze-and-Excitation (SE) Block \cite{hu2018squeeze}
         \item ConvNeXt Block \cite{liu2022convnet}
         \item BottleNeck Residual Block \cite{he2016deep}
         \item Inception-ResNetV2 Block \cite{szegedy2017inception}
     \end{itemize}
    Of these, the Inception-ResNetv2 block is the most effective. Its distinctive feature is the integration of Inception's parallel filter operations with residual connections, which serves to enhance feature learning. Furthermore, it employs a range of convolutional filter sizes, enabling the capture of diverse spatial features. Nevertheless, the results obtained in this experiment do not surpass those obtained by previous experiments utilising the original ResNetv1 block. 
    
\end{list}

\begin{table}[t]
\centering
\caption{Experimental modifications carried out during the design of the proposed architecture, for which performance is presented in Figure \ref{ablation_steps}.}
\label{ablation_steps_legend_2}
\footnotesize
\begin{tabular}{cclc}
\hline
Steps& ID & \multicolumn{1}{c}{Experimental Modifications}  & Params \\ 
\hline
\begin{tabular}[c]{@{}c@{}}Input\\ features\end{tabular} & \begin{tabular}[c]{@{}c@{}}I.1\\ I.2\\ I.3\\ I.4\\ I.5\\ I.6\end{tabular}                   & \begin{tabular}[c]{@{}l@{}}Discretization parameter: 0.05 cm\\ Discretization parameter: 0.2 cm\\ Discretization parameter: 0.5 cm\\ Features: $\vec{T}_c=(x_i, y_i, z_i)$\\ Features: $\vec{T}_c=(x_i, y_i, z_i, n_x, n_y, n_z)$\\ Features initialized at ones: $\vec{T}_c=(1)$ \end{tabular} &\begin{tabular}[c]{@{}l@{}}37.9M\\ 37.9M\\ 37.9M\\ 37.9M\\ 37.9M\\ 37.9M \end{tabular} 
\\ \hline
\begin{tabular}[c]{@{}c@{}}Global\\ design\end{tabular}  & \begin{tabular}[c]{@{}c@{}}G.1\\ G.2\\ G.3\\ G.4\\ G.5\\ G.6\\ G.7\\ G.8\end{tabular} & \begin{tabular}[c]{@{}l@{}}Cardinality: (2,3,4,6,2,2,2,2) $\rightarrow$ (2,2,2,2,2,2,2,2)\\ Cardinality: (2,3,4,6,2,2,2,2) $\rightarrow$ (1,1,1,1,1,1,1,1)\\ 4 skip connections $\rightarrow$ 3 skip connections\\ 3 skip connections  $\rightarrow$ 2 skip connections\\ Resnet Residual Block$\rightarrow$ BottleNeck Res. block \\ Resnet Residual Block$\rightarrow$ SE Block\\ Resnet Residual Block$\rightarrow$ ConvNext Res. Block\\ Resnet Residual Block$\rightarrow$ Inception-ResNetv2 Res. Block\end{tabular}  &\begin{tabular}[c]{@{}l@{}}21.7M\\ 11.6M\\ 11.5M\\ 11.4M\\ 14.7M\\ 11.5M \\ 66.4M \\ 2.9M \end{tabular} \\ 
\hline

\begin{tabular}[c]{@{}c@{}}Micro\\ design\end{tabular}   & \begin{tabular}[c]{@{}c@{}}M.1\\ M.2\\ M.3\\ M.4\\ M.5\\ M.6\\ M.7\end{tabular}       & \begin{tabular}[c]{@{}l@{}}Batch Normalization $\rightarrow$ Layer Normalization\\ Fewer activations\\ Kernel size 5 $\rightarrow$ Kernel size 3\\Kernel size 5 $\rightarrow$ Kernel size 9\\ Kernel size 5 $\rightarrow$ Kernel size 7\\ ReLu $\rightarrow$ GeLu \\ Drop Path Layer\end{tabular} &\begin{tabular}[c]{@{}l@{}}2.9M\\ 2.9M\\ 1.7M\\ 10M\\ 5.4M\\ 5.4M \\ 5.4M\end{tabular}                                               \\ \hline
\end{tabular}
\end{table}

\subsubsection{Micro Design}
\begin{list}{}{}
    \item{\textbf{Normalization}}: Likewise activation functions, it has become common practice to place operations that normalise the activations of each layer because of their multiple benefits in the learning process. As previously stated, two distinct approaches are examined in this phase of the investigation. Firstly, within the alternative experimental branch, which has its origins in node \textsl{G.2} and yielded the most favourable results using the original residual block of ResNetv1, the BatchNorm normalisation layer has been substituted with the Norm layer. However, this modification has caused a reduction in performance, with results declining from 78.3\% to 75.6\%. Secondly, the deployment of this layer within the configuration delineated by node \textsl{G.8} has led to an enhancement of 0.6\%. It may be reasonably deduced from the evidence presented that the normalisation layer in question performs more effectively after linear convolutions than after two-dimensional convolutions.
    

    \item{\textbf{Fewer activations}}: While activation functions after each convolutional layer is the most common approach, there are specific architectures and scenarios where selectively removing or moving them could make sense, like for computational efficiency or parameter reduction in mobile networks. However, this requires careful experimentation as haphazardly removing activations especially in deeper layers may impair performance by constraining the network's nonlinearity and representational capacity. In the alternative experimental branch (represented as the union of \textsl{G.2} and \textsl{M.1}), the elimination of the ReLu functions within the convolutional block of Resnetv1 is associated with a 3.8\% reduction in performance. However, in the main experimentation branch, the previous results are surpassed for the first time, with an increase of 1.4\% in comparison to the previous configuration and 0.5\% compared to the optimal result observed in the experiments. The global F1 score is 78.8\%.

    \item{\textbf{Kernel size}}: 
    Conventional CNNs typically utilise relatively small kernels. In contrast, Transformer-based architectures and CNN designs inspired by the Transformer often employ significantly larger kernels, with sizes beginning at 7x7 and beyond. This allows them to capture longer-range dependencies and broader context, contributing to their strong performance on various vision tasks. To explore the solutions proposed in Transformer-based architectures, a study of the convolution size has been conducted. Experiments \textsl{M.3}, \textsl{M.4} and \textsl{M.5}, depicted in Figure \ref{ablation_steps} and detailed in Table \ref{ablation_steps_legend_2}, demonstrate that a larger convolution window size leads to superior outcomes compared to the conventional convolution window sizes typically employed in convolutional networks (2, 3 or 5). Consequently, this variation results in an increase of 0.9\% in the F1 metric with respect to the previous optimal result.

    \item{\textbf{GeLu activation function.}} The ReLU has been the most popular activation function in deep learning due to its simplicity and efficiency. However, recent Transformer models have started using the GELU instead \cite{liu2021swin, liu2022convnet,alexey2020image}. GELU is a smoother variant of ReLU that can enable better training of deep networks by reducing issues like vanishing gradients and allowing useful gradients for negative input values. This activation function, experiment \textsl{M.6}, has resulted in an enhancement in performance relative to its predecessor, with an F1 metric of 80.5\%.

    \item{\textbf{Regularization technique.}} Dropout and drop-path are regularization techniques used in training deep neural networks to prevent overfitting and to improve generalization performance. Dropout randomly drops out (sets to zero) units in a layer during training, while drop-path randomly drops out entire paths or connections between layers. This prevents units or routes from becoming excessively reliant on one another, thereby ensuring their individual robustness. This results in a 1.5\% increase, which represents a final outcome of 82\% of the experimentation.
\end{list}

\section{Results and Discussions}
\label{section5}
This section offers a comprehensive performance evaluation of the proposed method, comparing it with established semantic segmentation and traversability estimation models from the literature. To ensure a fair comparison, we focus on accessible semantic segmentation models and methods originally designed for traversability estimation. The analysis covers a diverse range of approaches, including multi-layer perceptrons, traditional 2D convolutional networks, and cutting-edge sparse convolutions for efficient 3D data processing.
As emphasised throughout this paper, TE-NeXt is an architecture optimised for unstructured environments. However, it can also be regarded as a trade-off solution in urban environments. In order to facilitate a comprehensive and structured comparison, this analysis will be divided into two distinct Subsections, each of which will focus on a specific type of environment.

\subsection{Evaluation metrics}
The proposed method and other competitive state-of-the-art methods are evaluated based on the fundamental components of the confusion matrix: true positives (TP), true negatives (TN), false positives (FP), and false negatives (FN). Table \ref{Metrics} presents the equations for each calculated metric. Specifically, the accuracy metric measures the percentage of correctly classified instances and it is highly susceptible to bias due to the unequal distribution of instances across classes in an unbalanced dataset. In contrast, the F1 score metric provides a more reliable assessment as it is the harmonic mean between precision and recall, which aim to minimise false positives and false negatives, respectively. The Intersection over Union (IoU) is specifically computed as the ratio of the intersection area to the union area between the predicted and ground truth segmentation labels. The mean IoU (mIoU) is then computed by averaging the IoU values across all classes, ensuring each class contributes equally to the final performance metric. Finally, the true positive and true negative rate demonstrates the proportion of correctly classified true positive and true negative instances.

\renewcommand{\arraystretch}{1.5} 
\begin{table}[]
\centering
\caption{Performance metrics formulation.}
\label{Metrics}
\begin{tabular}{ll}
\hline
\multicolumn{1}{c}{Metric} & \multicolumn{1}{c}{Definition} \\ \hline

\multicolumn{1}{c}{Accuracy} &\multicolumn{1}{c}{$\frac{\text{TP} + \text{TN}}{\text{TP} + \text{TN} + \text{FP} + \text{FN}}$}
\\
\multicolumn{1}{c}{F1 score} & \multicolumn{1}{c}{$2 \cdot \frac{\text{Precision} \cdot \text{Recall}}{\text{Precision} + \text{Recall}}$}
\\
\multicolumn{1}{c}{mIoU} & \multicolumn{1}{c}{$\frac{1}{2} \left(\frac{\text{TP}}{\text{TP} + \text{FP} + \text{FN}} + \frac{\text{TN}}{\text{TN} + \text{FP} + \text{FN}}\right)$}  \\
\multicolumn{1}{c}{True Positive Rate} &  \multicolumn{1}{c}{$\frac{\text{TP}}{\text{TP} + \text{FN}}$}  \\
\multicolumn{1}{c}{True Negative Rate} & \multicolumn{1}{c}{$\frac{\text{TN}}{\text{TN} + \text{FP}}$} \\ \hline

\end{tabular}
\end{table}

\subsection{Performance in structured environments}

In the context of traversability estimation, our proposed solution, while effective, is outperformed by methods like SqueezeSegv3 and SalsaNext when assessed in a 2D domain using image projections of 3D data (rows 3 and 5 of Table \ref{urban_results} respectively). However, this 2D evaluation process introduces significant challenges and limitations. The projection from three-dimensional (3D) to two-dimensional (2D) space inevitably leads to information loss due to occlusions and depth ambiguities, where multiple points may map to a single pixel. When projecting segmentation results back to 3D space, issues such as one-to-many mapping and interpolation errors at object boundaries emerge, which result in a noticeable decrease in performance metrics \cite{alnaggar2021multi,cortinhal2020salsanext}.

Thus, these methods have also been evaluated in a 3D context in order to compare their performance under the same conditions as the other methods (rows 4 and 6 of Table \ref{urban_results} respectively). A decrease in their performance is observed, which confirms the limitations of 2D to 3D projections. To illustrate, SqueezeSegv3 demonstrates a decrease in its F1 score from 96.5\% in 2D to 95.8\% in 3D, whereas SalsaNext exhibits a reduction from 95.9\% to 94.9\%.
Consequently, while these methods may show superior results in 2D evaluations, they may not be optimal for critical traversability tasks in 3D real-world environments. In contrast, TE-NeXt demonstrates a solid balance between accuracy, precision, and reliability, achieving an F1 score of 96.0\% in structured environments. Thus, making it more suitable and robust for semantic segmentation in challenging, 3D real-world scenarios.

The comparison presented in Table \ref{urban_results} has been performed by training the semantic segmentation models from scratch, with the exception of Cyl3D, PVKD and P-SVM, which were adapted to the traversability problem in \cite{fusaro2023pyramidal} using the same approach. Furthermore, it has included additional traversability estimation methods, such as ProtoNet \cite{snell2017prototypical}, MPTI \cite{zhao2021few} and PBG \cite{bae2023self}. These metrics have been collected from the manuscripts of the aforementioned approaches, which have been subjected to the same testing conditions and code is not accessible.

\renewcommand{\arraystretch}{1.2} 
\begin{table*}[t]
\footnotesize
\caption{Traversability estimation performance of semantic segmentation methods in                      urban environments.}
\centering
\begin{tabular}{llccclcc}
\cline{1-8}
Urban envs. Sem. seg & \multicolumn{5}{c}{SemanticKITTI}\\ 
Methods             & \multicolumn{1}{l}{2D} & \multicolumn{1}{l}{3D} & \multicolumn{1}{l}{F1 score} & Accuracy & \multicolumn{1}{c}{mIOU}& \multicolumn{1}{l}{TPR} & \multicolumn{1}{l}{TNR}      \\ 
\cline{1-8}

PointNet \cite{qi2017pointnet}& &$\checkmark$ &93.5& 93.7& \multicolumn{1}{c}{88.1}& 96.5& 91.1  \\

PointNet++ \cite{qi2017pointnet++}& &$\checkmark$ & 83.3& 84.2& \multicolumn{1}{c}{72.4}& 81.1& 87.5 \\

SqueezeSegv3 \cite{xu2020squeezesegv3}        &$\checkmark$ & & \textbf{96.5}& \textbf{97.6}& \multicolumn{1}{c}{94.9}& \textbf{96.6}& 98.1 \\
SqueezeSegv3 \cite{xu2020squeezesegv3} & &$\checkmark$ & 95.8& 96.2& \multicolumn{1}{c}{92.7}& 95.7& 96.6 \\
 
SalsaNext \cite{cortinhal2020salsanext}&$\checkmark$ & & 95.9& 97.3& \multicolumn{1}{c}{\textbf{94.2}} & 95.0& \textbf{98.5}  \\ 
SalsaNext \cite{cortinhal2020salsanext}& &$\checkmark$ & 94.9& 95.5& \multicolumn{1}{c}{91.2} & 93.9& 96.7  \\
P-SVM \cite{fusaro2023pyramidal}& & $\checkmark$ & 89.2& 91.7& \multicolumn{1}{c}{83.9}& 89.0& 93.4   \\

RN++ \cite{milioto2019rangenet++}& & $\checkmark$ & 94.0& 95.4& \multicolumn{1}{c}{90.7} & 93.2& 96.8  \\

Cyl3D \cite{zhou2008cylinder3d}& & $\checkmark$ & 95.8& 96.8&\multicolumn{1}{c}{93.4} & 95.3& 97.7 \\

PVKD \cite{hou2022point}& & $\checkmark$ & 95.9& 96.9&\multicolumn{1}{c}{93.6} & 95.6& 97.7 \\ 

MinkU-Net \cite{choy20194d}& & $\checkmark$ & 93.7& 95.2&\multicolumn{1}{c}{90.0}&94.0& 93.8 \\
MinkU-Net \cite{choy20194d}+ ConvNeXt \cite{liu2022convnet} & & $\checkmark$ & 94.6& 95.2     &\multicolumn{1}{c}{90.8}& 97.6& 94.4 \\
\cline{1-8}
ProtoNet \cite{snell2017prototypical}& & $\checkmark$ & - & -&\multicolumn{1}{c}{80.4}&- & - \\

MPTI \cite{zhao2021few}& & $\checkmark$ & -& -&\multicolumn{1}{c}{85.9}&- & - \\

PBG \cite{bae2023self}& & $\checkmark$ & - & -&\multicolumn{1}{c}{84.1}& - & - \\
\cline{1-8}
TE-NeXt (ours)& & $\checkmark$ & 96.0& 96.3&\multicolumn{1}{c}{92.9}& 94.7& 97.7 \\ 
\cline{1-8}
\end{tabular}
\label{urban_results} 
\end{table*}

\subsection{Performance in natural environments}
 Table \ref{natural_results} presents the results of various semantic segmentation methods for traversability estimation in natural environments using the Rellis-3D dataset. The comparison presented in Table \ref{natural_results} is of particular relevance, given that natural environments represent the primary focus of the research. Unlike in urban environments, it is remarkable the good performance of the most classic method, PointNet, outperforming more modern methods in most of the metrics extracted.

The proposed architecture, TE-NeXt, exhibits the best overall performance, with an F1 score of 82.0\%, an accuracy of 88.3\% and an mIoU of 77.0\% for the non-risk and potentially dangerous areas for the robot's integrity. This outperforms other MinkU-Net-based approaches such as MinkU-Net+ConvNeXt, as well as techniques that transform the LiDAR input cloud into an image using projective geometry such as SqueezeSegV3 and SalsaNext.

\begin{table*}[t]
\footnotesize
\caption{Traversability estimation performance  of semantic segmentation methods in natural environments.}
\begin{center}
\begin{tabular}{llccclcc}
\cline{1-8}
Natural envs. Sem. seg & \multicolumn{5}{c}{Rellis-3D} \\ \cline{2-8}
Methods             & \multicolumn{1}{l}{ 2D}& \multicolumn{1}{l}{ 3D} & \multicolumn{1}{l}{F1 score} & Accuracy & \multicolumn{1}{c}{mIOU}& \multicolumn{1}{l}{TPR} & \multicolumn{1}{l}{TNR}\\ 
\cline{1-8}

PointNet \cite{qi2017pointnet}& &  $\checkmark$& 80.3& 87.2& \multicolumn{1}{c}{74.9} & 76.4& 92.8 \\
PointNet++ \cite{qi2017pointnet++}& & $\checkmark$& 58.9& 50.2& \multicolumn{1}{c}{32.1} & 45.6& 67.0\\

SqueezeSegv3 \cite{xu2020squeezesegv3}&  $\checkmark$& & 70.8& 87.9&\multicolumn{1}{c}{71.1} & 65.3& 95.1 \\
 SqueezeSegv3 \cite{xu2020squeezesegv3}& & $\checkmark$& 70.2 & 85.5 &\multicolumn{1}{c}{ 68.1} & 67.8 & 94.1  \\

SalsaNext \cite{cortinhal2020salsanext}& $\checkmark$ & & 73.5& 88.0&\multicolumn{1}{c}{73.3} & 68.9&\textbf{95.5} \\ 
 SalsaNext \cite{cortinhal2020salsanext}& & $\checkmark$&  70.9&  85.3&\multicolumn{1}{c}{ 70.5} &  65.4& 95.0 \\

MinkU-Net \cite{choy20194d}& & $\checkmark$& 75.8& 83.3&\multicolumn{1}{c}{69.1}   & 80.0& 84.9 \\
MinkU-Net \cite{choy20194d} + ConvNeXt \cite{liu2022convnet}& & $\checkmark$& 78.3& 85.9&\multicolumn{1}{c}{72.7} & 77.8& 89.8 \\ \cline{1-8}
nn Risk \cite{kiryo2017positive}& & $\checkmark$& -& -&\multicolumn{1}{c}{-}&70.4 & - \\

ScaTE \cite{seo2023scate}& & $\checkmark$& -& -&\multicolumn{1}{c}{-}&78.5 & - \\ \cline{1-8}

TE-NeXt (ours)& & $\checkmark$& \textbf{82.0}& \textbf{88.3}&  \multicolumn{1}{c}{\textbf{77.0}} & \textbf{81.3}& 91.8 \\ \cline{1-8}
\end{tabular}
\end{center}
\label{natural_results}
\end{table*}

In addition to a high mean intersection over union (mIoU) for traversability and non-traversability, the model achieves a true positive rate (TPR) of 81.3\%, indicating a robust capacity to accurately identify traversability areas. Concurrently, it exhibits a high true negative rate (TNR) of 91.8\%, which validates its efficacy in excluding non-traversability regions. In most of the metrics obtained, the TE-NeXt model exhibits superior performance, however, it is occasionally outperformed by methods evaluated in two-dimensional space, which demonstrate a considerable degree of variation in results between different metrics, such as TPR and TNR. This suggests that these methods are not consistent for this specific type of environment. Additionally, as previously stated, they demonstrate significantly lower performance when evaluated using the same 3D criteria as the other methods.

\begin{figure}[t]
\centering 
\includegraphics[width=0.75\textwidth, height=0.5\textwidth] {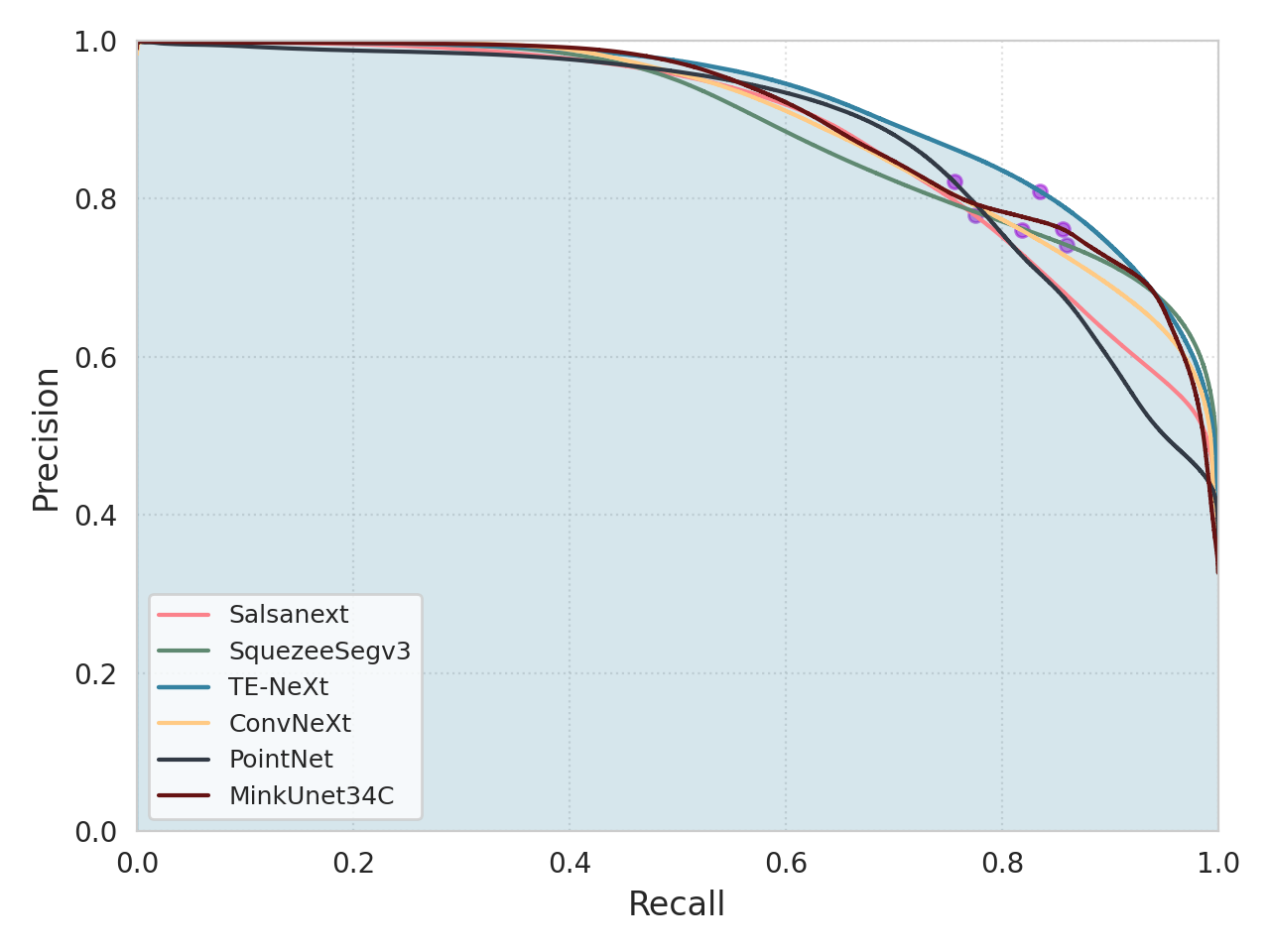}
\caption{Performance comparison between methods trained from scratch by means of Precision-Recall curve on Rellis-3D dataset.}
\label{pr-rellis} 
\end{figure}

In regard to methods dedicated to traversability estimation, such as nn Risk \cite{kiryo2017positive} and ScaTE \cite{seo2023scate}, which is at the state of the art as evidenced in Table \ref{sota}, TE-NeXt is superior in the single available metric, as the code is not accessible.

%

As the focus of this paper is on performance in unstructured environments, Figure \ref{pr-rellis} presents the precision-recall curves for the evaluated traversability estimation methods, providing a comprehensive visualization of their performance across different classification thresholds. This graphical representation offers valuable insights into the trade-offs between precision and recall for each method, particularly in identifying non-traversable areas. As observed in the figure, the proposed TE-NeXt method (blue line) exhibits superior performance. It maintains high precision across a wide range of recall values, suggesting robust identification of non-traversable areas with minimal false positives. This characteristic is crucial for traversability estimation in autonomous navigation, where maintaining high precision (minimizing false positives) is often prioritized to ensure safety. The area under the precision-recall curve (AUC-PR), provides a quantitative measure for the overall performance comparison. Based on the visual inspection of the curves, TE-NeXt presents a high edge in this regard, suggesting its potential as a robust solution for traversability estimation in autonomous navigation systems.

In order to complete the comparison, Table \ref{efficiency} provides a detailed quantitative analysis of the aforementioned methods in terms of efficiency. The results highlight significant differences in model size, memory requirements, and latency that can arise from architecture choices. Regarding to the model size, it is evident that TE-NeXt achieves one of its primary design goals: the reduction of the model's complexity without compromising its performance. 
Methods using 3D convolutions (MinkU-Net34C, MinkU-Net+ConvNeXt, TE-NeXt) tend to have lower training memory requirements compared to those using MLP or 2D convolutions, despite having more parameters in some cases. This suggests 3D convolutions may be more memory efficient during training.

\begin{table}
\caption{Comparison of semantic segmentation methods in terms of efficiency.}
\label{efficiency}
\footnotesize
\begin{center}
\begin{tabular}{lclclcl}
\cline{0-4}
Method's Efficiency   & Params & \begin{tabular}[c]{@{}l@{}}Type \end{tabular} & \begin{tabular}[c]{@{}l@{}}Training \\ Memory\end{tabular} & \begin{tabular}[c]{@{}l@{}}Inference \\ Latency\end{tabular} \\ \cline{0-4}
\multicolumn{1}{l|}{PointNet \cite{qi2017pointnet}}            & 3.5M & MLP    & 13GB        &   11 ms \\
\multicolumn{1}{l|}{PointNet++ \cite{qi2017pointnet++}}          & 0.9M & MLP  & 13.6GB        &   250 ms    \\
\multicolumn{1}{l|}{SqueezeSegv3 \cite{xu2020squeezesegv3}}        & 9.2M  & 2D    & 37.8GB   &   29 ms      \\
\multicolumn{1}{l|}{SalsaNext \cite{cortinhal2020salsanext}}           & 6.7M & 2D    & 11.2GB   &   28 ms                     \\
\multicolumn{1}{l|}{MinkU-Net \cite{choy20194d}}         & 37.9M & 3D   & 6.1GB    &   53ms     &  \\
\multicolumn{1}{l|}{MinkU-Net \cite{choy20194d}+ConvNeXt \cite{liu2022convnet}} & 66.4M & 3D  & 8.7GB    &   78 ms       \\
\multicolumn{1}{l|}{TE-NeXt (ours)}             & 5.4M & 3D    & 5.4GB    &   31 ms       \\ \cline{0-4}
\end{tabular}
\end{center}
\end{table}

\subsection{Qualitative Evaluation}

\begin{figure}[!ht]
    \centering
    \begin{subfigure}[b]{1\textwidth}
        \centering
        \includegraphics[width=1\textwidth, height = 0.55\textwidth]{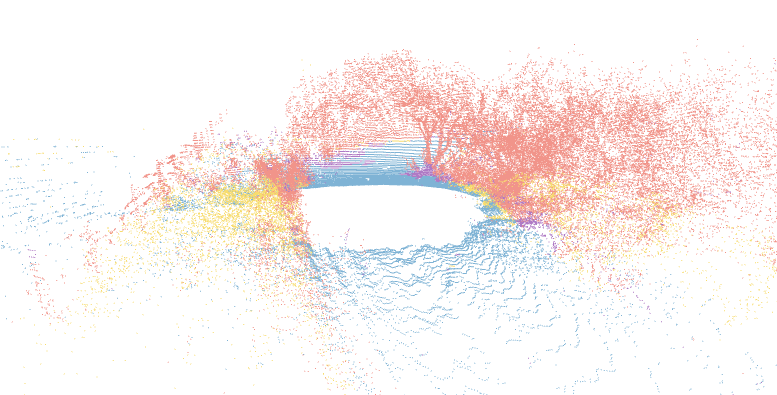}
        \caption{}
        \label{1a}
    \end{subfigure}
    \hfill
    \begin{subfigure}[b]{1\textwidth}
        \centering
        \includegraphics[width=1\textwidth, height = 0.55\textwidth]{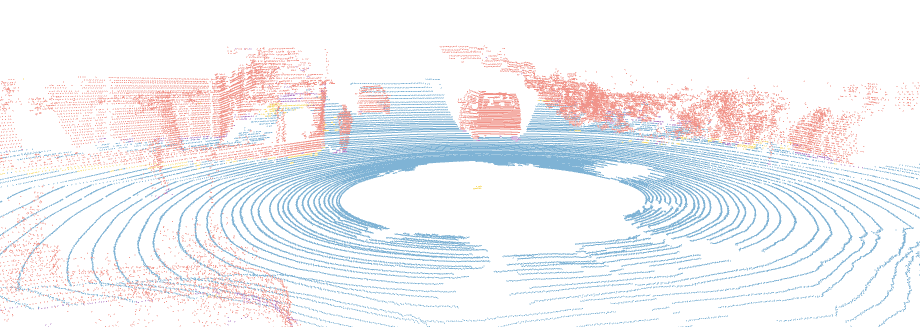}
        \caption{}
        \label{1b}
    \end{subfigure}
    \caption{Te-NeXt architecture inference visual representation. \textcolor{trav}{\rule[-0.07ex]{0.2cm}{1.6ex}}: true positives (TP). \textcolor{no_trav}{\rule[-0.07ex]{0.2cm}{1.6ex}}: true negatives (TN). \textcolor{fp}{\rule[-0.07ex]{0.2cm}{1.6ex}}: false positives (FP). \textcolor{fn}{\rule[-0.07ex]{0.2cm}{1.6ex}}: false negatives (FN). (a) Inference on a point cloud from the Rellis-3D dataset. (b) Inference on a point cloud from the SemanticKITTI dataset.}
    \label{fig:both}
\end{figure}

Figure \ref{1a} illustrates the outcomes of the proposed architecture in an unstructured environment. It can be observed that the points classified by the network as traversable but whose label is not (false positives) are located at the limits of the different geometries detected by the LiDAR. This error is attributed to the very nature of the algorithm itself, which discretises the environment into voxels. With regard to the false negatives, two scenarios emerge: firstly, those that present evidence in the right section of the image of the diffuse geometry of the environment. Secondly, the false negatives that appear in the left section of the image are indicative of an error in the labelling of the data, given that the traversability of this part of the environment is questionable.

Figure \ref{1b} shows the predictions of the proposed architecture in an urban environment. In this type of environment, as can be seen in the interpretation of results, the inference of the network is much more accurate, there are hardly any errors, and sources of error such as false positives occur for the same reasons mentioned above. 

\subsection{Fully Autonomous Navigation Framework}
In order to demonstrate the feasibility of the proposed architecture in dynamic, natural environments, an autonomous navigation system has been developed that employs the aforementioned traversability estimation approach. 
The incorporation of serial observations from sensors, such as GPS operating in real-time kinematic (RTK) mode, which enables centimetre-level accuracy, an inertial measuring unit (IMU), as well as encoders facilitating the odometry of the Husky A200 differential platform, allows for the calculation of the position and orientation of the robot relative to a known reference frame.

%

%

%

TE-NeXt enables local inference of the robot's surrounding environment, facilitating secure navigation to the next target point along a reasonably known global path. Its primary function is to help the robot navigate around unexpected obstacles, people, objects, or animals that may appear in the pre-calculated path. While traditional methods like clustering algorithms or occupancy grids can be employed for this purpose, they often lack of precision and robustness in contrast to deep learning-based approaches like TE-NeXt, especially in complex and unstructured environments where they may produce higher rates of false positives and negatives. The inferred point cloud serves as input to a trajectory generation algorithm in high-dimensional spaces, which is widely used in robotics, such as the Rapidly Exploring Random Trees (RRT) algorithm \cite{kuffner2000rrt}. Consequently, a series of three-dimensional coordinates is generated, representing a viable route between the robot's current location and the target point, as illustrated in Figure \ref{RRT_solution}.

In order to provide consistency for the navigation system, the control algorithm must now be designed to drive the robot from its current configuration, $(x_a, y_a, \alpha_a)$, to the goal position $(x_b, y_b, \beta_b)$, following correctly the generated trajectory. Applications such as \cite{stefek2020energy, malu2014kinematics} analyse the behaviour of a control system based on the Lyapunov stability criterion. This, in accordance with the kinematic model of differential drive mobile robot, gives rise Equations \ref{v} and \ref{w}.
\begin{equation}
\label{v}
v = \left\| \mathbf{x} \right\|_2  k_v \cos(\theta_e)
\end{equation}
\begin{equation}
\label{w}
\omega = k_\omega \cos(\theta_e) \sin(\theta_e) + k_\omega \theta_e
\end{equation}

where:
\begin{itemize}
    \item \( v \) is the linear velocity.
    \item \( \omega \) is the angular velocity.
    \item \( k_v \) and \( k_\omega \) are positive constants.
    \item $\left\| \mathbf{x} \right\|_2 = \sqrt{\sum_{i=1}^{n} x_i^2}$ is the norm 2 error in position.
    \item  \(\theta_e\)  is the orientation error.
\end{itemize}

\begin{figure}
\centering 
\includegraphics[width=0.7\textwidth, height=0.4\textwidth]{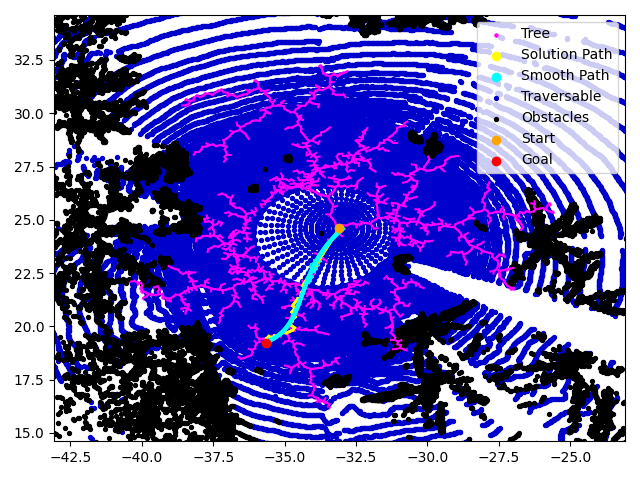}
\caption{The multiple assumptions made by the RRT algorithm and demonstration that one of these assumptions is capable of satisfying the proposed constraints.}
\label{RRT_solution} 
\end{figure}

It can be seen from Figure \ref{overview_autonomous_system} that the aforementioned tools, when working together, develop an autonomous navigation system that is capable of adapting dynamically to changes in the environment. This ensures that the paths generated are not only possible but also safe.
This adaptability is crucial for real-world applications, where conditions can change rapidly. The complete result can be viewed here\footnote{\url{https://www.youtube.com/watch?v=ZZIoeuBu60I}}. The video demonstrates the generation and tracking of trajectories at two levels: i) at the local level, where the hypotheses generated by the RRT algorithm are shown, and the path that satisfies the constraints is highlighted; and ii) at the global level, where a geo-referenced trajectory with the respective waypoints is displayed. Additionally, the video presents the inference of 1100 LiDAR scans processed by TE-NeXt, which evaluates the feasibility of the calculated path along the entire trajectory.

\begin{figure}
\centering 
\includegraphics[width=1\textwidth, height=0.6\textwidth]{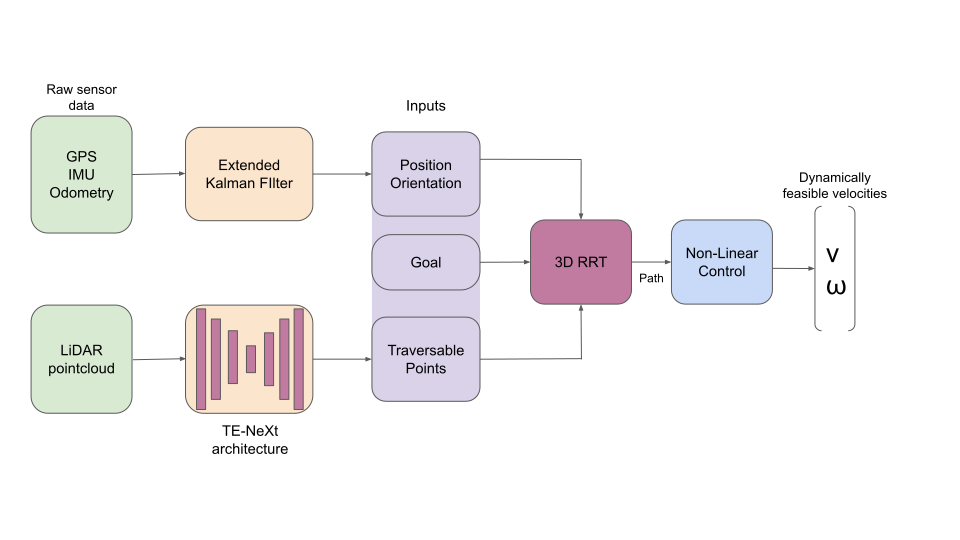}
\caption{Integration of the proposed architecture into a customised autonomous navigation system.}
\label{overview_autonomous_system}
\end{figure}

\subsection{Implementation details}
All the experiments have been tested on an Intel  Core\texttrademark{} i9-10900X, 20 $\times$ 3.70GHz, 128GB RAM platform with a NVIDIA RTX 3090 with 24GB VRAM graphic card. The neural network model is implemented using the Minkowski Engine \cite{choy20194d}, Pytorch \cite{paszke2019pytorch} and it has been integrated into ROS NOETIC. Reproducibility is ensured by a standardized report altogether with the dataset, which are publicly available at Github\footnote{\url{https://github.com/ARVCUMH/te-next}}.



\section{Conclusion}
\label{section6}

This document presents TE-NeXt, a technique for estimating traversability based on a sparse three-dimensional convolutional neural network. The architecture is an encoder-decoder configuration inspired by Mink-Unet, which makes a significant contribution through its comprehensive study of a tailored convolutional block. This block demonstrates greater abstraction capacity in natural environments than classical ResNet blocks. Emphasising the importance of the proposed residual block, which allows to employ large 3D kernel sizes to model long-range dependencies combined with linear convolutions that reduce computational cost, allow cross-channel mixing, adjust dimensions for skip connections, and provide learnable feature fusion. The proposed architecture exploits an efficient volumetric representation through MinkU-Net \cite{ronneberger2015u} and the deep learning capabilities of techniques that have shown robustness in semantic segmentation problems, such as vision Transformers \cite{liu2021swin} and ConvNeXt \cite{liu2022convnet}.

Furthermore, the comparison results presented in this manuscript validate the effectiveness of the proposed approach, TE-NeXt, for traversability estimation in highly natural environments, achieving results of 82\% on the F1 metric, composed of the harmonic mean of precision and recall, and 77\% on the mIoU metric, among others. It thus demonstrates superior performance compared to the current state-of-the-art methods, offering enhanced stability in terms of performance. Nevertheless, further research and improvements to existing methods are required to cope with the increasing complexity of natural scenes. Additionally, the network demonstrates comparable results to the most advanced semantic segmentation methods in urban environments.

Future work will include the addition of visual information through a LiDAR-camera calibration, which will allow visual features to be introduced into the sparse tensors. Possibly with the aim of mitigating the main problem of supervised learning, the annotation process, we will investigate the exposure of a hybrid method that allows continuous estimation of traversability.

\section{Acknowledgments}
This work has been funded by the ValgrAI Foundation, Valencian Graduate School and Research Network of Artificial Intelligence through a predoctoral grant. In addition, this publication is part of the project TED2021-130901B-I00, funded by MCIN/AEI/10.13039/501100011033 and by the European Union ``NextGenerationE''/PRTR'' and of the projects PROMETEO/2021/075 funded by the Generalitat Valenciana. The authors acknowledge Funda\c{c}\~{a}o para a Ci\^encia e a Tecnologia (FCT) for its financial support via the project LAETA Base Funding (DOI:10.54499/UIDB/50022/2020).



\bibliographystyle{elsarticle-num}
\bibliography{main}
\end{document}